\documentclass[letterpaper, 10 pt, conference]{ieeeconf}  % Comment this line out if you need a4paper

\IEEEoverridecommandlockouts                              % This command is only needed if 
\usepackage{graphicx}
\usepackage{amsmath} % assumes amsmath package installed
\usepackage{amssymb}  % assumes amsmath package installed

\title{\LARGE \bf
SWTrack: Multiple Hypothesis Sliding Window \\ 3D Multi-Object Tracking 
}

\author{Sandro Papais$^{1}$, Robert Ren$^{1}$, and Steven Waslander$^{1}$% <-this % stops a space
\thanks{$^{1}$University of Toronto Robotics Institute, University of Toronto, Toronto, M5S 1A4, Canada. {\tt\small sandro.papais@mail.utoronto.ca}
}%
}

\begin{document}

\maketitle
\thispagestyle{empty}
\pagestyle{empty}

%%%%%%%%%%%%%%%%%%%%%%%%%%%%%%%%%%%%%%%%%%%%%%%%%%%%%%%%%%%%%%%%%%%%%%%%%%%%%%%%
\begin{abstract}

Modern robotic systems are required to operate in dense dynamic environments, requiring highly accurate real-time track identification and estimation. For 3D multi-object tracking, recent approaches process a single measurement frame recursively with greedy association and are prone to errors in ambiguous association decisions. Our method, Sliding Window Tracker (SWTrack), yields more accurate association and state estimation by batch processing many frames of sensor data while being capable of running online in real-time. The most probable track associations are identified by evaluating all possible track hypotheses across the temporal sliding window. A novel graph optimization approach is formulated to solve the multidimensional assignment problem with lifted graph edges introduced to account for missed detections and graph sparsity enforced to retain real-time efficiency. We evaluate our SWTrack implementation on the NuScenes autonomous driving dataset to demonstrate improved tracking performance.
\end{abstract}

%%%%%%%%%%%%%%%%%%%%%%%%%%%%%%%%%%%%%%%%%%%%%%%%%%%%%%%%%%%%%%%%%%%%%%%%%%%%%%%%
\section{INTRODUCTION}

3D multi-object tracking (MOT) is a fundamental component of robotic systems operating in dynamic environments, such as autonomous vehicles. Accurate tracking performance is essential to predict the future motion of nearby objects and perform path planning. MOT performs a central role by temporally aggregating perception information from sensors into histories that describe object motion in the environment. 

Tracking performance is affected by several sources of uncertainty. First, object detections from modern deep object detectors are often noisy and include many false positives and false negatives. Second, the motion of objects is uncertain, it can vary for different objects and change over time. Finally, the number of objects to be tracked is unknown and changes as objects enter and leave the field of view of the robot. These issues create ambiguities in what new observations to associate with existing tracks, when to initiate new tracks, and when to delete old tracks.

\begin{figure}[!htb] % TODO: Replace image with more clear example
\centering
\includegraphics[width=3.4in]{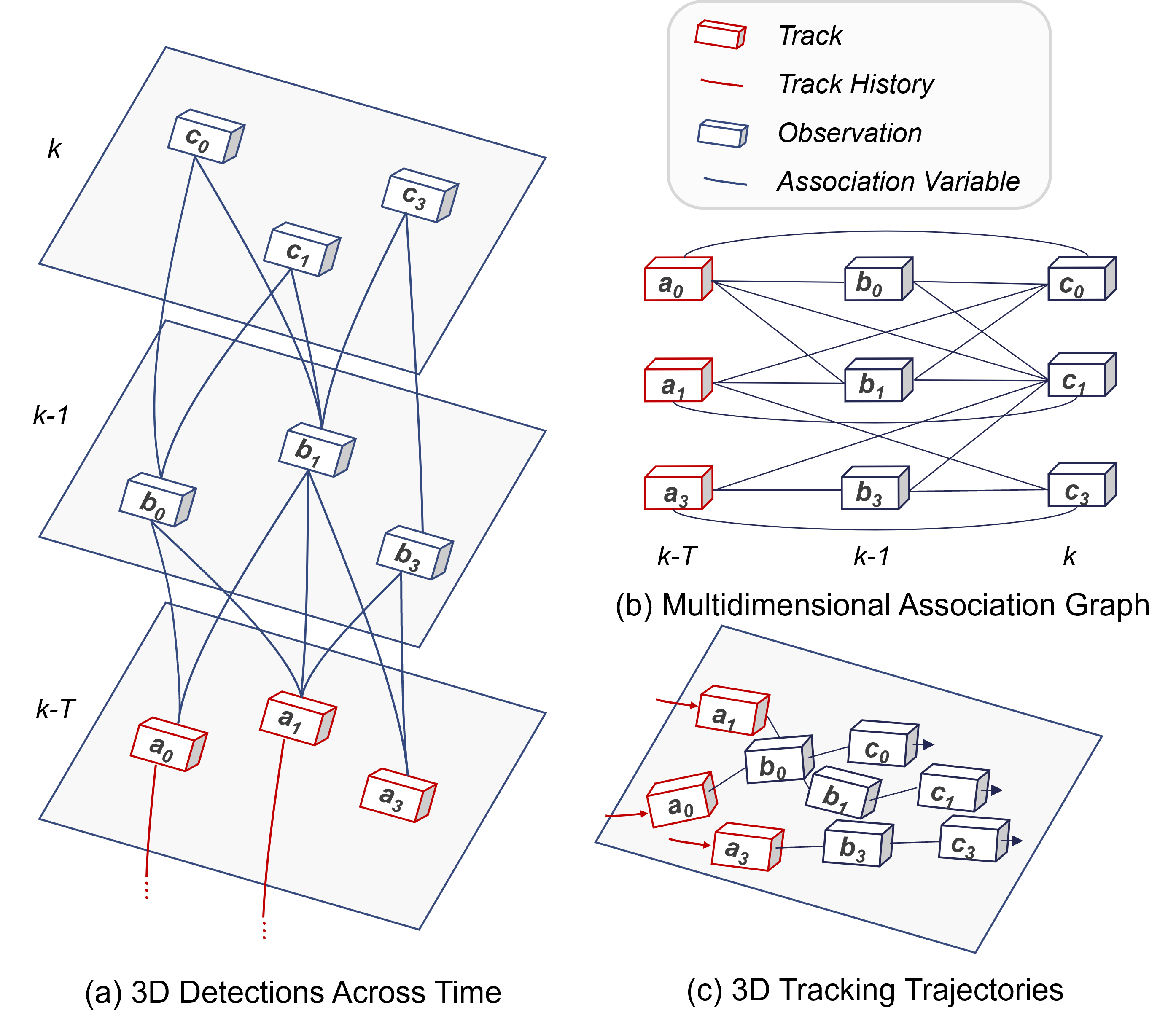}
\caption{An example illustration for multiple hypothesis tracking with noisy observations from 3 objects. Observations of objects are shown as blue boxes, association decision variables are blue lines, tracked objects are red boxes, and track histories are red lines. Our sliding window tracker solves an association problem over many frames and can rectify decisions made in past frames with more recent information. The association is solved by optimizing over a multidimensional graph structure to obtain 3D trajectories.}
\label{fig:illustration}
\end{figure}

Most recent 3D multi-object tracking algorithms rely on single-frame recursive filtering \cite{yinCenterbased3dObject2021, chenVoxelNeXtFullySparse2023, kimEagerMOT3DMultiObject2021, chiuProbabilistic3DMultiModal2021}. These methods solve data association of current observations to previous tracks. The track filters are updated with the associated measurements and association decisions cannot be revised when new information becomes available in the future. The single-frame approach is sufficient in simple tracking scenarios with few objects, limited occlusion, and low measurement noise. However, in more challenging operational environments such as densely populated scenes, ambiguities arise that cannot be resolved in the current frame and lead to tracking errors.

In order to improve tracking performance we developed a sliding window tracker that solves the tracking problem over a temporal window of observations instead of a single frame at a time (see Figure~\ref{fig:illustration}). We formulate the multidimensional assignment problem over multiple frames as a weighted graph optimization. The graph is assembled iteratively online and unlikely graph edges are pruned to reduce the feasible solution space. Track hypotheses are formed by enumerating all possible traversals of the remaining graph edges and limiting to the top $N$ hypothesis candidates. Graph pruning and top $N$ hypotheses strategies are used to address the exponential complexity of hypothesis growth and reduce the optimization problem to a more tractable size.

The likelihood of each track hypothesis is estimated by the log-likelihood ratio of observations forming a track to observations being noise or another object. A likelihood ratio is derived that combines multiple independent sources of information to support hypothesis selection: kinematic likelihood, detection confidence, and LIDAR or camera measurement similarity. The globally optimal tracking solution is approximated by finding the highest likelihood track association hypotheses and track states for our reduced problem. 

The key contributions of this work are to (1) develop a novel multidimensional graph optimization formulation of multiple hypothesis sliding window tracking for mobile robotics applications, (2) create a sparse graph structure with lifted edges that can be efficiently assembled and solved online, and (3) derive a log-likelihood ratio for hypothesis scoring using kinematic likelihood, feature embedding similarity, and detection confidence. We verify our proposed method more accurately estimates object trajectories in challenging tracking scenarios for autonomous driving compared to the typical single-hypothesis tracking approaches. 

%%%%%%%%%%%%%%%%%%%%%%%%%%%%%%%%%%%%%%%%%%%%%%%%%%%%%%%%%%%%%%%%%%%%%%%%%%%%%%%%

\section{RELATED WORK}

In 3D multi-object tracking, the standard approach uses a 3D object detection algorithm to generate observations from sensors such as LIDARs and cameras \cite{weng3DMultiObjectTracking2020}, known as tracking-by-detection. A distance or similarity metric is computed for each pair of object detections and tracked objects predicted in the current frame. The association problem is a weighted bipartite graph or linear assignment problem and is typically solved by the greedy nearest neighbor method or a Hungarian algorithm variation \cite{kuhn1955hungarian, crouseImplementing2DRectangular2016}. The matching solution is used to update the track positions in the current frame. Tracks are initiated after multiple consecutive matches are made and deleted after no association is made for a given duration. 

The standard 3D tracking approach was detailed in AB3DMOT \cite{weng3DMultiObjectTracking2020} which used a 3D intersection over union association metric. In CenterPoint \cite{yinCenterbased3dObject2021} the object detector was trained to regress velocity estimates for each bounding box observation, the velocity measurements are used for tracking with a simple Euclidean distance metric for the association. VoxelNeXt \cite{chenVoxelNeXtFullySparse2023} builds on CenterPoint by tracking using query voxels or object key points instead of tracking using predicted object centers.

Several other 3D tracking methods propose to incorporate camera information to help with tracking. EagerMOT \cite{kimEagerMOT3DMultiObject2021} added a multi-stage association approach using 2D object detections from cameras and 3D object detections from LIDAR to maintain 2D tracks and 3D tracks in parallel, where tracks can be updated by a 3D object detection, 2D object detection, or both. ProbabilisticMOT \cite{chiuProbabilistic3DMultiModal2021} tracks 3D objects using a learned distance metric based on the fusion of 3D LIDAR features, 2D camera features, and 3D Mahalanobis distances. CAMO-MOT \cite{wangCAMOMOTCombinedAppearanceMotion2023} builds on ProbabilisticMOT by further learning model occlusions. % TODO: switch probmot papers?

An alternative approach to tracking involves joint detection and tracking. AlphaTrack \cite{zengCrossModal3DObject2021a} jointly trains a LIDAR object detection branch and camera embedding branch, using position and visual features for tracking. SimTrack \cite{luoExploringSimple3D2021} performs detection and tracking using pairs of point clouds to regression motion for a simple learned tracker. PFTrack \cite{pangStandingFutureSpatioTemporal2023} is an end-to-end trained multi-camera 3D framework using transformers to represent tracked instances over time as object queries. 3D MODT \cite{kini3DMODTAttentionGuidedAffinities2023} performs joint detection and tracking in 3D point cloud data also using a transformer encoder. While these methods show promise they usually do not perform as well as tracking-by-detection methods. 

The multidimensional assignment problem has not been extensively explored in the literature for robotics applications. However, multiple-hypothesis tracking (MHT) was formulated by Reid in 1979 \cite{reidAlgorithmTrackingMultiple1979} to improve tracking performance in cluttered environments. Subsequent formulations expanded on this method with additional approaches and more efficient solutions \cite{blackmanMultipleHypothesisTracking2004}. While MHT is a well-known concept in multi-target tracking, it has seen limited application in mobile robotics. Our approach extends the multiple hypothesis tracking idea by mapping it to a multidimensional graph optimization problem that can be efficiently solved online for robotics applications and is simpler to implement than Lagrangian relaxation multidimensional assignment methods \cite{deb1997generalized, poore1994multidimensional}. 

%%%%%%%%%%%%%%%%%%%%%%%%%%%%%%%%%%%%%%%%%%%%%%%%%%%%%%%%%%%%%%%%%%%%%%%%%%%%%%%%

\section{METHODOLOGY}

\begin{figure*}[!htb]
\centering
\includegraphics[width=6.8in]{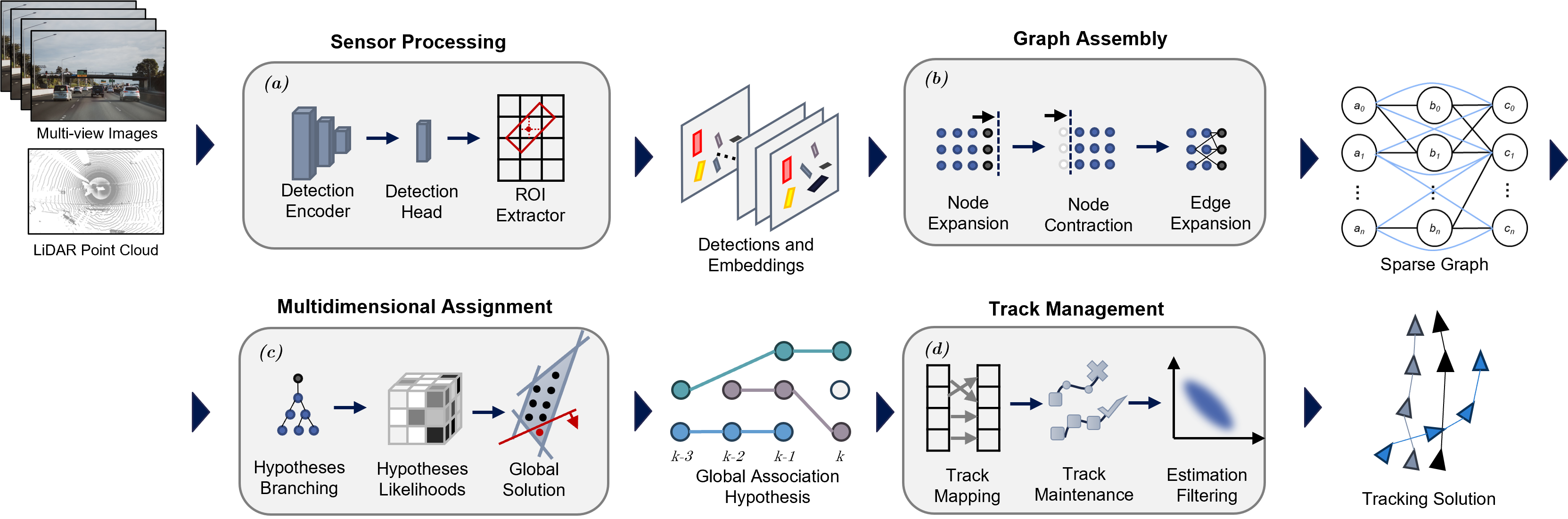}
\caption{Sliding window tracker architecture. (a) Measurement data is passed through a 3D encoder to extract feature embeddings and generate detections. (b) Detected objects are used to expand the sparse graph nodes and find new edges with lifted edges shown as blue lines. (c) Track hypotheses and likelihoods are drawn from the graph and solved by linear programming. (d) Tracks are assigned IDs and filtered to obtain the tracking solution. }
\label{fig:arch}
\end{figure*} % TODO: replace lidar/images

The sliding window tracker is made up of four primary components as detailed in Figure~\ref{fig:arch}. LIDAR and Camera sensor data are processed into feature embedding and detections. A sparsely connected multidimensional association graph is assembled from detections with the addition of lifted long-range edges. The association graph is mapped to a track hypotheses selection optimization problem that can be efficiently solved. The global association solution is used to assign track identification numbers to observations, filter track states, delete old tracks, and initiate new tracks.

\subsection{Preliminaries}

The sliding window tracking goal is to estimate the states $\mathbf{x}_i=\mathbf{x}_{i,k-T:k}=\left\{\mathbf{x}_{i,k-T}, \ldots, \mathbf{x}_{i,k-1}, \mathbf{x}_{i,k}\right\}$ at discrete times $k-T$ to $k$ for each tracked object $i$ given the associated noisy measurements $\mathbf{y}=\mathbf{y}_{i,k-T:k}=\left\{\mathbf{y}_{i,k-T}, \ldots, \mathbf{y}_{i,k-1}, \mathbf{y}_{i,k}\right\}$. The object states are represented by $ \mathbf{x}_{i,k} = \begin{bmatrix} \mathbf{r}_{i,k}^\intercal & \boldsymbol \theta_k^\intercal & \mathbf{v}_{i,k}^\intercal & \mathbf{s}_{i,k}^\intercal \end{bmatrix}^\intercal$,  where $\mathbf{r}_{i,k} \in \mathbb{R}^3$ is the object center position in the global frame, $\boldsymbol \theta_{i,k} = \begin{bmatrix} 0 & 0 & \theta_{i,k} \end{bmatrix}^\intercal$ is the orientation expressed using only yaw, $\mathbf{v}_{i,k} \in \mathbb{R}^3$ is the velocity, and $\mathbf{s}_{i,k} = \begin{bmatrix} l_{i,k} & w_{i,k} & h_{i,k} \end{bmatrix}^\intercal$ is the bounding box dimensions. 

To estimate object trajectories we first find the most probable measurements, $\mathbf{y}_i$, for a given tracked object $i$. We evaluate the track hypothesis probability or track score using the standard likelihood ratio \cite{sittler1964optimal,blackmanDesignAnalysisModern1999a} between the target hypothesis and the null hypothesis as
\begin{equation}
    \Lambda_i = \frac{p(H_1|\mathcal{D}_i)}{p(H_0|\mathcal{D}_i)} = \frac{p\left(\mathcal{D}_i | H_1\right) p\left(H_1\right)}{p\left(\mathcal{D}_i | H_0\right) p\left(H_0\right)},
\end{equation}
where the target hypothesis, $H_1$, is the measurements, $\mathbf y_i$, belongs to the same object. For convenience, the log-likelihood ratio, $\lambda_i = \log(\Lambda_i)$, is used as the track score and can be converted to track probabilities $p(H_1|\mathcal{D}_i) = e^{\lambda_i}/(1+e^{\lambda_i})$. 

\subsection{Online Sparse Graph Assembly}

The association objective is to find the set of track hypotheses with the maximum likelihood ratio without reusing the same observation more than once. The multidimensional assignment problem at time $k$ over the past $T$ frames can be expressed as a dense directed acyclic graph, $G^{dense}_{k-T:k} = (V;E^{d})$, with $V$ nodes being the detections across the sliding window, and $E^{d}$ edges are all possible associations of detections across the sliding window. A key distinction in our work is that the graph is defined to include skip connections or lifted edges between non-adjacent frames in order to allow for the association of objects through occlusion or missed detections. 

The dense multidimensional association graph has a graph order and graph size given by
\begin{equation}
    |V| = \sum_{i=k-T}^k N_i, \quad |E^{d}| = \sum_{i=k-T}^k\sum_{j=i}^k N_i N_j,
\end{equation}
where $N_i$ is the number of detections in frame $i$. The set of possible track hypotheses, $Z$, on the dense graph is given by
\begin{equation}
    |Z| = \prod_{i=k-T}^k (N_i+1) - \sum_{i=k-T}^k (N_i) - 1.
\end{equation}
The track hypotheses can take on $N_i+1$ states in each frame $i$ including the additional null state that represents a skipped connection. Hypotheses that do not contain at least two detections are not removed. Due to the exponential number of hypotheses to be evaluated, $\mathcal{O}(N^T)$, the solution is intractable for a large number of frames as it is an NP-hard combinatorial optimization problem \cite{gareyComputersIntractabilityGuide1979}.

To approximate the multidimensional assignment we define a sparse graph, $G^{sparse}_{k-T:k} = (V;E^{s})$, by retaining the nodes from the dense graph and removing low probability edges. For edge deletion, two strategies are implemented: class-based and distance-based. Class-based edge deletion is used to remove edges between observations from different predicted classes (car, truck, pedestrian, etc.). Distance-based edge deletion removes pairs of objects that exceed a class-wise 99.9 percentile of the velocity error distribution. Since edges can span multiple time steps the distance limit is $d_\mathrm{lim} = v_\mathrm{lim} \delta t$ with an upper bound to avoid distant solutions.

% \begin{figure}[!htb]
% \centering
% \includegraphics[width=3.0in]{figs/graphs.png}
% \caption{Comparison of the multidimensional association graphs with a sliding window of size 3 and $n$ detections per frame. The dense graph (left), $G^{dense}_{k-T:k}$, and sparse graph (right), $G^{sparse}_{k-T:k}$, are shown with lifted edges spanning multiple frames as blue lines and adjacent edges as black lines. Detection nodes are shown as black circles and track nodes are shown in red circles. An edge connects two nodes while a hypothesis is a set of edges connecting multiple nodes.}
% \label{fig:graphs}
% \end{figure}

The graph is assembled online by reusing the previous graph and modifying it. The graph is contracted at frame $k-T-1$ by removing the oldest nodes. Nodes are added at frame $k-T$ to include dormant tracks older than frame $k-T$ up to a threshold $k-T-T_{\mathrm{old}}$. The graph is expanded at frame $k$ by adding nodes for the newest detections received. For each new node added, edges are added to every node in other frames. For each new edge, distance-based and class-based deletion rules are also applied.

\subsection{Multidimensional Assignment}

Using the association graph we define the multidimensional assignment problem in a compact matrix form as the integer program given by
\begin{align} \begin{split} \label{eqn:ip}
\arg\max_{\mathbf z_k} \quad & \mathbf c^{\top}_k \mathbf z_k, \\
\text {subject to} \quad & \mathbf{A}_k \mathbf{z}_k \leq \mathbf b_k, \quad \mathbf z_k \in \left\{0, 1 \right\}^{p \times 1},
\end{split} \end{align}
where $\mathbf z_k \in \left\{0, 1 \right\}^{p \times 1}$ is the binary decision variable that represents the acceptance of $p$ track hypotheses, $\mathbf{c}_k \in \mathbb{R}^{p \times 1}$ is the cost vector with the likelihood of each track hypothesis, $\mathbf A_k \in \left\{0, 1 \right\}^{q \times p}$ is the binary constraint matrix with $q$ constraints, and $\mathbf {b}_k = \mathbf{1}^{q \times 1}$ is the constraint vector.

The mapping between the multidimensional assignment problem and association graph is defined by the hypothesis map, $\mathbf{z}_{\text{map}_{i,j}} \in \mathbb{Z}^{p \times T}$. Each map entry, $\mathbf{z}_{\text{map}_{i,j}}$, contains the graph node index for the $i$th hypothesis at the $j$th frame with a value from 1 to $N_j$ or 0 to indicate a skipped connection. The track hypotheses map $\mathbf{z}_{\text{map}}$ contains all the possible traversals of the graph $G^{sparse}_{k-T:k}$. 

Hypotheses tree management follows a detection-oriented approach. The latest observations form the root of the track hypothesis trees. At each new frame, the hypotheses map is expanded by branching each hypothesis into new ones via new edges in the graph. During tree expansion, the past hypotheses are combined with a new root node for each detection. During tree contraction, one frame of nodes is removed. The hypotheses vector is then contracted again by taking the $M$ best track hypotheses from each family of hypotheses originating from a given node at time $k$. 

The number of hypotheses before and after enforcing $M$-best hypotheses approximation are 
\begin{equation}
    p^{\mathrm{d}} = \prod_{i=k-T}^k\left(N_i+1\right) - \sum_{i=k-T}^k\left(N_i\right)-1, ~~~~  p^{\mathrm{s}} = \sum_{i=1}^{N_k} M.
\end{equation}
Therefore, the decision variables are reduced from exponential in the number of frames, $\mathcal O (N^T)$, to linear in the number of max hypotheses, $\mathcal O(NM)$, leading to a significant reduction in hypotheses count and solution time. Graph sparsity further helps reduce the memory and computation cost to build the optimization problem by removing unlikely hypotheses before even computing them.

% \begin{figure}[!htb]
% \centering
% \includegraphics[width=3.4in]{figs/hyp_tree.png}
% \caption{Hypotheses tree management strategy follows a detection-oriented approach. The latest observations form the root of the tree. During tree expansion, the past hypotheses are combined with a new root node for each detection. During tree contraction, one frame of nodes in removed, and any remaining hypotheses below the top $m$-best scores for a given root node are removed.}
% \label{fig:hyp_tree}
% \end{figure}

The constraint $\mathbf{A}_k \mathbf{z}_k \leq \mathbf b_k$ ensures that each detection cannot be used more than once in the solution. The number of constraints is equal to the number of nodes in the graph, $q = \sum_{i=k-T}^k N_i$. The matrix $\mathbf A_k$ is extremely sparse and the number of nonzero elements is given by $ \mathrm{nnz}(\mathbf{A}_k) \approx TN^T$ for the case where we have the same number of detections in every frame, $N$. This leads to a sparsity $(\mathbf A_k) \geq 0.99$ for typical tracking problems ($N \geq 100$) that allows this formulation to be solved efficiently by commercial solvers.

% The matrix $\mathbf A_k$ is extremely sparse and the number of nonzero elements is given by $ \mathrm{nnz}(\mathbf{A}_k) \approx TN^T$ for the case where we have the same number of detections in every frame, $N$. This leads to
% \begin{align}
%     \operatorname{sparsity}(\mathbf{A}_k) &= \frac{\operatorname{numel}(\mathbf{A}_k)-\operatorname{nnz}(\mathbf{A}_k)}{\operatorname{numel}(\mathbf{A}_k)} \\
%     &\approx \frac{TN^{T+1} - TN^T}{TN^{T+1}} = 1 - \frac{1}{N}.
% \end{align}
% Therefore, the sparsity $(\mathbf A_k) \geq 0.99$ for typical tracking problems ($N \geq 100$). This allows this formulation to be solved efficiently by commercial solvers.

The integer program in Equation~\ref{eqn:ip} generally cannot be solved in polynomial time. However, since the $\mathbf A_k$ is a totally unimodular matrix and $\mathbf b_k$ is integral then the linear programming (LP) relaxation has an integral optimum when an optimum exists. Therefore, we can solve an LP relaxation
\begin{align}
\arg\max_{\mathbf z_k} \quad & \mathbf c^{\top}_k \mathbf z_k, \\
\text {subject to} \quad & \mathbf{A}_k \mathbf{z}_k \leq \mathbf b_k, \quad \mathbf 0 \leq \mathbf z_k \leq \mathbf 1.
\end{align}
online with a guarantee of a solution in polynomial time.

% TODO: reduction back to single frame single hypothesis

\subsection{Track Hypothesis Likelihood}
\begin{table*}[!htb]
\centering
\caption{AMOTA evaluation results on NuScenes validation set.}
\label{tab:nuscenes_class_results}
\begin{tabular}{c|c|c|c|c|c|c|c|c|c}
\hline
Tracker        & Detector               & overall &  bicycle & bus & car & motorcycle & pedestrian & trailer & truck \\ \hline\hline
CenterPoint \cite{yinCenterbased3dObject2021}    & CenterPoint \cite{yinCenterbased3dObject2021}            & 63.8 & 40.9 & 79.9  & 82.9 & 54.6 & 73.6 & 48.8 & 65.2 \\ 
ProbalisticMOT \cite{chiuProbabilistic3DMultiModal2021} & CenterPoint \cite{yinCenterbased3dObject2021}            & 61.4 & 38.7 & 79.1 & 78.0 & 52.8 & 69.8 & 49.4 & 62.2     \\ 
SWTrack        & CenterPoint \cite{yinCenterbased3dObject2021}            & 65.6 & 48.0 & 85.1 & 77.9 & 59.6 & 80.4 & 41.4 & 67.0 \\ \hline
CenterPoint \cite{yinCenterbased3dObject2021}    & BEVFusion  \cite{liuBEVFusionMultiTaskMultiSensor2023}            & 70.4 & 66.9 & 85.1 & 84.7 & 76.3 & 59.3 & 49.6 & 71.6    \\ 
SWTrack        & BEVFusion \cite{liuBEVFusionMultiTaskMultiSensor2023}              & 71.6 & 64.8 & 82.8 & 83.2 & 72.9 & 74.9 & 48.7 & 70.0 \\ \hline
\end{tabular}
\end{table*}

The log-likelihood ratio for hypothesis $i$ is computed as
\begin{equation}
    \mathbf{c}_i = \log \left( \frac{p\left(\mathcal{D}_i | H_1\right) p\left(H_1\right)}{p\left(\mathcal{D}_i | H_0\right) p\left(H_0\right)} \right).
\end{equation}
We consider multiple sources of independent data to support each hypothesis, $p\left(\mathcal{D}_i\right) = p\left(\mathcal{D}_i^\mathrm{k},\mathcal{D}_i^\mathrm{s},\mathcal{D}_i^\mathrm{c}\right)$, such that
\begin{equation}
        \mathbf{c}_i = \log \left( 
        \frac{p(\mathcal{D}_i^\mathrm{k}|H_1)}{p(\mathcal{D}_i^\mathrm{k} | H_0)}  
        \frac{p(\mathcal{D}_i^\mathrm{s}|H_1)}{p(\mathcal{D}_i^\mathrm{s} | H_0)}
        \frac{p(\mathcal{D}_i^\mathrm{c}|H_1)}{p(\mathcal{D}_i^\mathrm{c} | H_0)}
        \frac{p(H_1)}{p(H_0)}
        \right).
\end{equation}
Therefore the log-likelihood ratio between a given track hypothesis, $i$, and the null hypothesis is modeled with four factors: the prior, kinematic likelihood of the trajectory, detection signal confidence, and embedding similarity. 
\begin{equation}
    \mathbf{c}_i = \mathbf{c}_{i}^{\text{prior}} + \mathbf{c}_i^{\text{kin}} + \mathbf{c}_i^{\text{conf}} + \mathbf{c}_i^{\text{sim}}.
\end{equation}
The prior log-likelihood ratio is a constant and can be omitted. The kinematic log-likelihood ratio for a given track hypothesis $i$ at frame $k$ is computed as
\begin{equation}
    \mathbf c_{i}^{\text{kin}} = \log\left(\frac{p(\mathcal{D}_i^\mathrm{kin}|H_1)}{p(\mathcal{D}_i^\mathrm{kin}|H_0)}\right) 
    = \log\left(\frac{p(\mathbf y_{i,0:k}|\mathbf x_{i,0:k})}{p(\mathbf y_{i,0:k}|H_0)}\right),
\end{equation}
where $\mathbf y_{i,k}$ is the most recent measurement, $k$, for track hypothesis $i$ and $H_0$ is the null hypothesis that the measurement $\mathbf y_{i,k}$ is not part of the track. We assume a linear Gaussian model with a constant velocity motion model
\begin{equation}
    \mathbf x_{i,k} = \mathbf A_{i,k-1} \mathbf x_{i,k}, \quad \mathbf y_{i,k} = \mathbf C_{i,k} \mathbf x_{i,k} + \mathbf n_{i,k},
\end{equation}
with process noise, $\mathbf{w}_k \sim \mathcal{N}\left(\mathbf{0}, \mathrm{Q}_k\right)$, and measurement noise, $\mathbf{n}_k \sim \mathcal{N}\left(\mathbf{0}, \mathrm{R}_k\right)$. Uncorrelated noise is assumed to factor the likelihood ratio and iteratively compute it as
\begin{equation}
    \mathbf c_{i}^{\text{kin}} = \log\left(\prod_{j=0}^k \frac{p(\mathbf y_{i,j}|\mathbf x_{i,j})}{p(\mathbf y_{i,j}|H_0)}\right).
\end{equation}
The kinematic likelihood of the measurement $(i,k)$ at time $k$ belonging to hypothesis $i$ is modeled as a Gaussian distribution and likelihood of a false detection is a uniform distribution over the measurement volume $V$
\begin{equation}
    \Delta c_{i,k}^{\text{kin}} = \log\left(\frac{\mathcal{N}\left(\mathbf{y}_{i,k} | \check{\mathbf{x}}_{i,k}, \check{\mathbf{P}}_{i,k}\right)}{U(V)}\right),
\end{equation}
\begin{equation}
    \Delta c_{i,k}^{\text{kin}} \propto \frac{1}{2} \log \left|\check{\mathbf{P}}_{i,k}\right|-\frac{\mathbf d_{i,k}^2}{2},
\end{equation}
where constant terms are omitted, $\mathbf d_{i,k}^2 = (\mathbf y_{i,k}-\check{\mathbf{x}}_{i,k})^\top \check{\mathbf{P}}_{i,k}^{-1} (\mathbf y_{i,k}-\check{\mathbf{x}}_{i,k})$ is the squared Mahalonobis distance, $\check{\mathbf{x}}_{i,k}$ is the predicted state of track hypothesis, and $\check{\mathbf{P}}_{i,k}$ is the predicted covariance. A Kalman filter is run for each hypothesis to predict the states and covariance of each object.

The appearance similarity log-likelihood ratio for a given track hypothesis $i$ at frame $k$ is defined as 
\begin{equation}
    \mathbf c_{i}^{\text{sim}} = 
     \log \left(\frac{p(\mathbf X_{i,k}|\mathbf X_{i,0:k-1})}{p(\mathbf X_{i,k}|H_0)}\right),
\end{equation}
where $\mathbf X_{i,k}$ is the sensor data at frame $k$ corresponding to object $i$. Since the geometry and appearance of objects may change over time we model the similarity across neighboring pairs of frames and iteratively update it online
\begin{equation}
    \mathbf c_{i}^{\text{sim}} =
     \log \left( \prod_{j=0}^k \frac{p(\mathbf X_{i,j}|\mathbf X_{i,j-1})}{p(\mathbf X_{i,j}|H_0)}\right).
\end{equation}
To compute the similarity ratio we compute an embedding for each object $i$ at time $k$ by bilinear interpolation as
\begin{equation}
    \mathbf h_{i,k} = \mathrm{BilinInterp}(f(\mathbf X_{i,k}), (u_{i,k}, v_{i,k})),
\end{equation}
where $f(\mathbf X_{i,k})$ is the encoder feature map and $(u_{i,k}, v_{i,k})$ are the normalized coordinates of the object location in the feature map. The pairwise similarity metric is given by the cosine similarity as $\mathrm{sim}(\mathbf h_1,\mathbf h_2) = \mathbf{h_1}^{\top} \mathbf{h_2} /\|\mathbf{h_1}\|\|\mathbf{h_2}\|$. For pairs of embeddings of objects across frames, we can compute the normalized probability using the softmax function
\begin{equation}
    \Delta c_{i,k}^{\text{sim}} \propto
     \log \left( \frac{\exp(\mathrm{sim}(\mathbf h_{i,k},\mathbf h_{i,k-1}))}{ \sum_{j \neq i} \exp(\mathrm{sim}(\mathbf h_{i,k},\mathbf h_{j,k-1}))} \right),
\end{equation}
where $p(\mathbf X_{i,j}|H_0)$ is assumed to be constant and omitted. The classifier confidence scores, $f_{i}$, are used to compute the confidence likelihood ratio given the presence of a detection.
\begin{equation}
    \mathbf c_{i}^{\text{conf}} = \log\left(\frac{p(\mathcal{D}_i^\mathrm{conf}|H_1)}{p(\mathcal{D}_i^\mathrm{conf}|H_0)}\right) =
     \log \left(\frac{p(\mathbf f_{i}, \mathbf y_{i}|H_1)}{p(\mathbf f_{i}, \mathbf y_{i}|H_0)}\right).
\end{equation}
The joint probability of the confidence scores, $\mathbf f_{i}$, and a detection occurring, $\mathbf y_{i}$, can be expanded for each frame 
\begin{equation}
    \Delta c_{i,k}^{\text{conf}} = 
     \log \left(\frac{p(\mathbf f_{i}|\mathbf y_{i},H_1)p(\mathbf y_{i}|H_1)}{p(\mathbf f_{i}|\mathbf y_{i},H_0)p(\mathbf y_{i}|H_0)}\right) \propto \log\left(\mathbf f_{i} \right).
\end{equation}
If no detection is present in the track hypothesis and the frame is skipped then 
\begin{equation}
    \Delta c_{i,k}^{\text{conf}} = 
     \log\left(\frac{1-P_{\mathrm{D}}}{1-P_{\mathrm{FA}}}\right), \quad  \Delta c_{i,k}^{\text{kin}} =  \Delta c_{i,k}^{\text{sim}} = 0,
\end{equation}
where $P_{\mathrm{D}}$ is the probability of detection given a ground truth object taken as the detector true positive ratio and $P_{\mathrm{FA}}$  is the probability of false alarm which is the detector false negative ratio. The log-likelihood ratio is computed recursively at each new frame $k$ for each hypothesis $i$ as
\begin{equation}
    c_{k,i} = c_{k,i} + \Delta c_{k,i}^{\text{kin}} + \Delta c_{k,i}^{\text{conf}} + \Delta c_{k,i}^{\text{sim}}.
\end{equation}

\subsection{Track Management}

Track management is responsible for mapping the tracking solution over the sliding window to the track decisions made in the past. In order to maintain real-time performance and causality in the tracker, the updated solution to previous time steps is not updated in the official track solution. At each time step the best decision is output by mapping the global hypothesis to track-detection pairs in the current frame. 

The tracker outputs the current tracks as a list of bounding boxes with a unique track ID. Since the sliding window tracker is able to revise old association decisions it is possible that a track solution may have multiple past track IDs, in this case, the most recent track ID is taken for the solution.

While the tracker only outputs the current tracks for the purpose of evaluation, one additional benefit of this method is that it can also improve the accuracy of past tracks. In robotics applications, the entire history is helpful since it can be used for the prediction of future motion and is the most accurate tracking information.

We follow standard rules for track maintenance by initiating a track once it has been observed at least twice. If the track has not been seen for 4 frames it is deleted. Track states are filtered using a Kalman filter for the track states and the box dimensions are averaged over past observations.
% TODO: double check 

%%%%%%%%%%%%%%%%%%%%%%%%%%%%%%%%%%%%%%%%%%%%%%%%%%%%%%%%%%%%%%%%%%%%%%%%%%%%%%%%
\begin{figure*}[!t]
\centering
\includegraphics[width=6.0in]{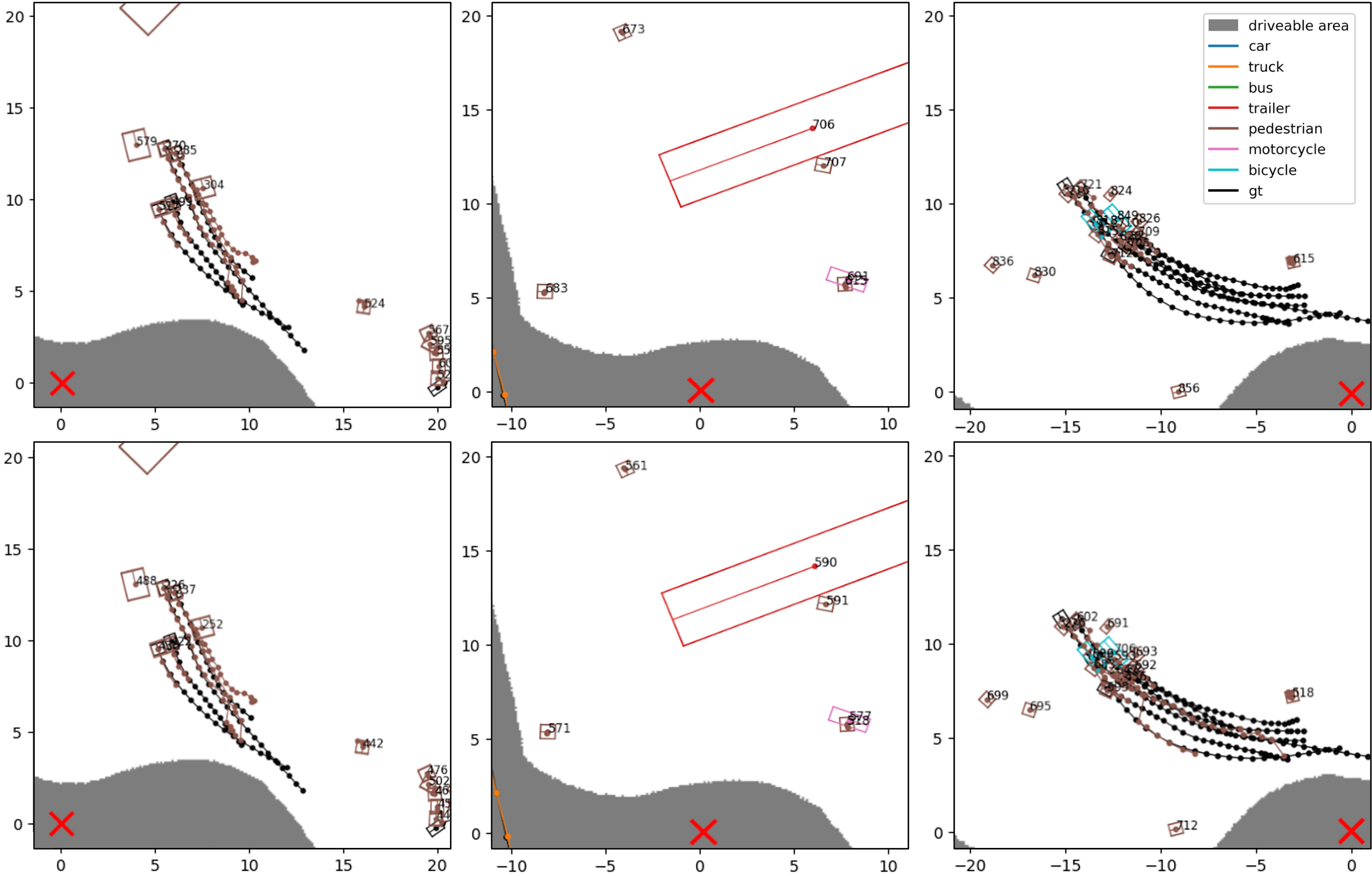}
\caption{Qualitative tracking comparison with a horizon of two frames (top) and four frames (bottom) on the NuScenes validation set. In the first column, a group of 5 pedestrians are being tracked. In the second column, 2 seconds later, the tracks are lost due to occlusion. In the third column, 2 seconds later, the objects reappear and the sliding window tracker is able to recover the full track history while the single frame tracker creates new tracks.}
\label{fig:qual}
\end{figure*}

\section{EXPERIMENTAL RESULTS}

We evaluate our method on the NuScenes \cite{caesarNuScenesMultimodalDataset2020a} dataset. The NuScenes validation set contains 6019 labeled samples at 2 Hz. The samples are split into continuous scenes of 20 seconds long. The tracker is initialized at the start of each scene. Our proposed method is tested using embeddings and detections from two 3D object detection methods. The first detector is CenterPoint \cite{yinCenterbased3dObject2021} using LIDAR data as inputs. The second detector is BEVFusion \cite{liuBEVFusionMultiTaskMultiSensor2023} which encodes LIDAR and multi-view camera inputs. We only compare tracking results against methods with the same input detections for a fair comparison with other state-of-the-art methods. We evaluate using various sliding window lengths from 0.5 to 2.5 seconds and the maximum hypotheses per node set to 200. The Gurobi solver \cite{gurobi} is used in all experiments.

Performance is evaluated using Average Multi-Object Tracking Accuracy (AMOTA) \cite{bernardinEvaluatingMultipleObject2008}. The MOTA metric is
\begin{equation}
    \mathrm{MOTA}=1-\frac{\mathrm{FP}+\mathrm{FN}+\mathrm{IDS}}{\mathrm{GT}},
\end{equation}
where $\mathrm{GT}$ is the number of ground truth objects, $\mathrm{FP}$ is a false positive object track, $\mathrm{FN}$ is a false negative object track, and $\mathrm{IDS}$ is an identity switch meaning the object is in the correct position but the association was incorrect. The metric is averaged over a set of $L$ discrete recall values or track score confidence thresholds, $r \in\left\{\frac{1}{L}, \frac{2}{L}, \cdots, 1\right\}$.
\begin{equation}
    \resizebox{.87\hsize}{!}{$\mathrm{AMOTA}=\frac{1}{L} \sum_{r}\left(1-\frac{\mathrm{FP}_r+\mathrm{FN}_r+\mathrm{IDS}_r-(1-r) \mathrm{GT}}{r \cdot \mathrm{GT}}\right)$}.
\end{equation}

We benchmark our method, SWTrack, using the CenterPoint detections against the ProbablisticMOT \cite{chiuProbabilistic3DMultiObject2020} LIDAR variant and CenterPoint\cite{yinCenterbased3dObject2021} trackers. We also benchmark against the CenterPoint tracker using the improved BEVFusion detections. As shown in Table~\ref{tab:nuscenes_class_results} and Table~\ref{tab:nuscenes_results_summary} our method outperforms both previous tracking methods when using both detector variations. This shows that our method is able to correctly reason over multiple frames to improve the tracking performance over single-frame tracking approaches. Our method in particular performs much better for tracking difficult classes with less predictable motion such as pedestrians, bicycles, and motorcycles.

\begin{table}[!htb]
\centering
\caption{Evaluation results on NuScenes validation set.}
\label{tab:nuscenes_results_summary}
\begin{tabular}{c|c|c|c|c|c}
\hline
Tracker        & Detector               & AMOTA & FP    & FN    & IDS \\ \hline\hline
CenterPoint    & CenterPoint            & 63.8  & 18612 & 22928 & 760 \\ 
SWTrack        & CenterPoint            & 65.6  & 11280 & 23035 & 926 \\ \hline
CenterPoint    & BEVFusion              & 70.4  &  14014  &  21386     &  744   \\ 
SWTrack        & BEVFusion              & 71.6  & 12986 & 19621 & 849 \\ \hline
\end{tabular}
\end{table}

To evaluate the improvements of our method we study the ablation of the sliding window length parameter $T$ using the BEVFusion detector, shown in Table~\ref{tab:ablate_window}. When using a sliding window length of 2 frames, the sliding window tracker reduces to a simple single frame tracker and shows similar performance to the CenterPoint tracker as expected. As the number of frames is increased there is an increase in performance of 0.5, 0.6, and 0.2 AMOTA for each frame number increase. In addition, the number of track identity switches (IDS), is reduced by 336 when switching from a tracking horizon of 2 to 4, which illustrates an improved ability to associate observations. When tested with a horizon of more than 5 frames the performance improvements show diminishing improvements on NuScenes while incurring significant increases in computational overhead. 

Despite the complexity growth as the sliding window horizon expands, we find that for $T = 4$ frames, we can operate reliably at 0.1 s latency on an i7-9800X CPU. When benchmarking the runtime it was found that the computational bottleneck was during the assembly of the optimization problem rather than solving it. This suggests it is possible to further speed up the implementation by improving the code vectorization and parallel computation.

\begin{table}[!htb]
\centering
\caption{Ablation on the number of frames batch processed in the sliding window tracker.}
\label{tab:ablate_window}
\begin{tabular}{c|c|c|c|c}
\hline
SW Length & AMOTA & FP    & FN    & IDS  \\ \hline\hline
2 (0.5 s)  &   70.5  &  11298  & 20129   &  1185 \\ 
3 (1.5 s)   &   71.0  &  17031  & 21733   &  994  \\ 
4 (2.0 s)   &   71.6  &  12986  & 19621   &  849 \\ 
5 (2.5 s)   &   71.8 &  12735 &  18943 & 801  \\ \hline
\end{tabular}
\end{table}

A study of the qualitative performance of the sliding window tracker has shown three main improvements: occlusion handling, false positive reduction, and track length. First, the tracker is able to maintain tracks through occlusion better by increasing the receptive temporal range of association as shown in Figure~\ref{fig:qual}. Single-frame trackers like CenterPoint are typically restricted to maintaining tracks for 3 frames without a new detection before deleting them, while we can effectively reason about longer tracks and score them in relation to other hypotheses.

The sliding window tracker also exhibits significantly fewer false positive detections. Since the log-likelihood ratio cost function incorporates detection confidence, similarity, and distance metrics it is able to efficiently filter out spurious detections. Tracks are also maintained much longer on average due to the ability to avoid switching track IDs with low-quality detections.

% Table: Offline vs Online

%%%%%%%%%%%%%%%%%%%%%%%%%%%%%%%%%%%%%%%%%%%%%%%%%%%%%%%%%%%%%%%%%%%%%%%%%%%%%%%%

\section{CONCLUSION}

Sliding Window Tracker solves the multidimensional assignment problem by efficiently constructing a sparse graph with lifted edges to model complex association connections. This graph is used to approximate the multiple hypothesis tracking solution and generate an association of current detections to existing tracks. This method is well tailored to 3D robotics perception in difficult dynamic environments since it can better model difficult associations and easily incorporate multiple forms of available data derived from modern sensors into track likelihoods. Our approach demonstrates an updated formulation to apply multiple hypothesis tracking to modern robotics applications such as autonomous driving.

%%%%%%%%%%%%%%%%%%%%%%%%%%%%%%%%%%%%%%%%%%%%%%%%%%%%%%%%%%%%%%%%%%%%%%%%%%%%%%%%

\addtolength{\textheight}{-10cm}   % This command serves to balance the column lengths

%\section*{APPENDIX}


%\section*{ACKNOWLEDGMENT}

%The preferred spelling of the word “acknowledgment” in America is without an “e” after the “g”. Avoid the stilted expression, “One of us (R. B. G.) thanks . . .”  Instead, try “R. B. G. thanks”. Put sponsor acknowledgments in the unnumbered footnote on the first page.

%%%%%%%%%%%%%%%%%%%%%%%%%%%%%%%%%%%%%%%%%%%%%%%%%%%%%%%%%%%%%%%%%%%%%%%%%%%%%%%%

\bibliographystyle{./IEEEtran}
\bibliography{IEEEabrv,3DMOT}

\begin{thebibliography}{10}
\providecommand{\url}[1]{#1}
\csname url@rmstyle\endcsname
\providecommand{\newblock}{\relax}
\providecommand{\bibinfo}[2]{#2}
\providecommand\BIBentrySTDinterwordspacing{\spaceskip=0pt\relax}
\providecommand\BIBentryALTinterwordstretchfactor{4}
\providecommand\BIBentryALTinterwordspacing{\spaceskip=\fontdimen2\font plus
\BIBentryALTinterwordstretchfactor\fontdimen3\font minus
  \fontdimen4\font\relax}
\providecommand\BIBforeignlanguage[2]{{%
\expandafter\ifx\csname l@#1\endcsname\relax
\typeout{** WARNING: IEEEtran.bst: No hyphenation pattern has been}%
\typeout{** loaded for the language `#1'. Using the pattern for}%
\typeout{** the default language instead.}%
\else
\language=\csname l@#1\endcsname
\fi
#2}}

\bibitem{yinCenterbased3dObject2021}
T.~Yin, X.~Zhou, and P.~Krahenbuhl, ``Center-based 3d object detection and
  tracking,'' in \emph{Proceedings of the {{IEEE}}/{{CVF Conference}} on
  {{Computer Vision}} and {{Pattern Recognition}}}, 2021, pp. 11\,784--11\,793.

\bibitem{chenVoxelNeXtFullySparse2023}
Y.~Chen, J.~Liu, X.~Zhang, X.~Qi, and J.~Jia, ``{{VoxelNeXt}}: {{Fully Sparse
  VoxelNet}} for {{3D Object Detection}} and {{Tracking}},'' in \emph{2023
  {{IEEE}}/{{CVF Conference}} on {{Computer Vision}} and {{Pattern
  Recognition}} ({{CVPR}})}, 2023, pp. 21\,674--21\,683.

\bibitem{kimEagerMOT3DMultiObject2021}
A.~Kim, A.~O{\v s}ep, and L.~{Leal-Taix{\'e}}, ``{{EagerMOT}}: {{3D
  Multi-Object Tracking}} via {{Sensor Fusion}},'' in \emph{2021 {{IEEE
  International Conference}} on {{Robotics}} and {{Automation}} ({{ICRA}})},
  May 2021, pp. 11\,315--11\,321.

\bibitem{chiuProbabilistic3DMultiModal2021}
H.-K. Chiu, J.~Li, R.~Ambru{\c s}, and J.~Bohg, ``Probabilistic {{3D
  Multi-Modal}}, {{Multi-Object Tracking}} for {{Autonomous Driving}},'' in
  \emph{2021 {{IEEE International Conference}} on {{Robotics}} and
  {{Automation}} ({{ICRA}})}, May 2021, pp. 14\,227--14\,233.

\bibitem{weng3DMultiObjectTracking2020}
X.~Weng, J.~Wang, D.~Held, and K.~Kitani, ``{{3D Multi-Object Tracking}}: {{A
  Baseline}} and {{New Evaluation Metrics}},'' in \emph{2020 {{IEEE}}/{{RSJ
  International Conference}} on {{Intelligent Robots}} and {{Systems}}
  ({{IROS}})}, Oct. 2020, pp. 10\,359--10\,366.

\bibitem{kuhn1955hungarian}
H.~W. Kuhn, ``The hungarian method for the assignment problem,'' \emph{Naval
  research logistics quarterly}, vol.~2, no. 1-2, pp. 83--97, 1955.

\bibitem{crouseImplementing2DRectangular2016}
D.~F. Crouse, ``On implementing {{2D}} rectangular assignment algorithms,''
  \emph{IEEE Transactions on Aerospace and Electronic Systems}, vol.~52, no.~4,
  pp. 1679--1696, Aug. 2016.

\bibitem{wangCAMOMOTCombinedAppearanceMotion2023}
L.~Wang, X.~Zhang, W.~Qin, X.~Li, J.~Gao, L.~Yang, Z.~Li, J.~Li, L.~Zhu,
  H.~Wang, and H.~Liu, ``{{CAMO-MOT}}: {{Combined Appearance-Motion
  Optimization}} for {{3D Multi-Object Tracking With Camera-LiDAR Fusion}},''
  \emph{IEEE Transactions on Intelligent Transportation Systems}, pp. 1--16,
  2023.

\bibitem{zengCrossModal3DObject2021a}
Y.~Zeng, C.~Ma, Z.~Ming, Z.~Fan, and X.~Yang, ``Cross-{{Modal 3D Object
  Detection}} and {{Tracking}} for {{Auto-Driving}},'' in \emph{2021
  {{IEEE}}/{{RSJ International Conference}} on {{Intelligent Robots}} and
  {{Systems}} ({{IROS}})}, Sept. 2021, p.~8.

\bibitem{luoExploringSimple3D2021}
C.~Luo, X.~Yang, and A.~Yuille, ``Exploring {{Simple 3D Multi-Object Tracking}}
  for {{Autonomous Driving}},'' in \emph{Proceedings of the {{IEEE}}/{{CVF
  International Conference}} on {{Computer Vision}}}, 2021, pp.
  10\,488--10\,497.

\bibitem{pangStandingFutureSpatioTemporal2023}
Z.~Pang, J.~Li, P.~Tokmakov, D.~Chen, S.~Zagoruyko, and Y.-X. Wang, ``Standing
  {{Between Past}} and {{Future}}: {{Spatio-Temporal Modeling}} for
  {{Multi-Camera 3D Multi-Object Tracking}},'' in \emph{2023 {{IEEE}}/{{CVF
  Conference}} on {{Computer Vision}} and {{Pattern Recognition}}
  ({{CVPR}})}.\hskip 1em plus 0.5em minus 0.4em\relax {Vancouver, BC, Canada}:
  {IEEE}, June 2023, pp. 17\,928--17\,938.

\bibitem{kini3DMODTAttentionGuidedAffinities2023}
J.~Kini, A.~Mian, and M.~Shah, ``{{3DMODT}}: {{Attention-Guided Affinities}}
  for {{Joint Detection}} \& {{Tracking}} in {{3D Point Clouds}},'' in
  \emph{2023 {{IEEE International Conference}} on {{Robotics}} and
  {{Automation}} ({{ICRA}})}, May 2023, pp. 841--848.

\bibitem{reidAlgorithmTrackingMultiple1979}
D.~Reid, ``An algorithm for tracking multiple targets,'' \emph{IEEE
  Transactions on Automatic Control}, vol.~24, no.~6, pp. 843--854, Dec. 1979.

\bibitem{blackmanMultipleHypothesisTracking2004}
S.~Blackman, ``Multiple hypothesis tracking for multiple target tracking,''
  \emph{IEEE Aerospace and Electronic Systems Magazine}, vol.~19, no.~1, pp.
  5--18, Jan. 2004.

\bibitem{deb1997generalized}
S.~Deb, M.~Yeddanapudi, K.~Pattipati, and Y.~Bar-Shalom, ``A generalized sd
  assignment algorithm for multisensor-multitarget state estimation,''
  \emph{IEEE Transactions on Aerospace and Electronic systems}, vol.~33, no.~2,
  pp. 523--538, 1997.

\bibitem{poore1994multidimensional}
A.~B. Poore, ``Multidimensional assignment formulation of data association
  problems arising from multitarget and multisensor tracking,''
  \emph{Computational Optimization and Applications}, vol.~3, no.~1, pp.
  27--57, 1994.

\bibitem{sittler1964optimal}
R.~W. Sittler, ``An optimal data association problem in surveillance theory,''
  \emph{IEEE transactions on military electronics}, vol.~8, no.~2, pp.
  125--139, 1964.

\bibitem{blackmanDesignAnalysisModern1999a}
S.~S. Blackman and R.~Popoli, \emph{Design and {{Analysis}} of {{Modern
  Tracking Systems}}}.\hskip 1em plus 0.5em minus 0.4em\relax {Artech House},
  1999.

\bibitem{gareyComputersIntractabilityGuide1979}
M.~R. Garey and D.~S. Johnson, \emph{Computers and {{Intractability}}: {{A
  Guide}} to the {{Theory}} of {{NP-Completeness}}}.\hskip 1em plus 0.5em minus
  0.4em\relax {W. H. Freeman}, 1979.

\bibitem{liuBEVFusionMultiTaskMultiSensor2023}
Z.~Liu, H.~Tang, A.~Amini, X.~Yang, H.~Mao, D.~L. Rus, and S.~Han,
  ``{{BEVFusion}}: {{Multi-Task Multi-Sensor Fusion}} with {{Unified
  Bird}}'s-{{Eye View Representation}},'' in \emph{2023 {{IEEE International
  Conference}} on {{Robotics}} and {{Automation}} ({{ICRA}})}, May 2023, pp.
  2774--2781.

\bibitem{caesarNuScenesMultimodalDataset2020a}
H.~Caesar, V.~Bankiti, A.~H. Lang, S.~Vora, V.~E. Liong, Q.~Xu, A.~Krishnan,
  Y.~Pan, G.~Baldan, and O.~Beijbom, ``{{nuScenes}}: {{A Multimodal Dataset}}
  for {{Autonomous Driving}},'' in \emph{2020 {{IEEE}}/{{CVF Conference}} on
  {{Computer Vision}} and {{Pattern Recognition}} ({{CVPR}})}.\hskip 1em plus
  0.5em minus 0.4em\relax {Seattle, WA, USA}: {IEEE}, June 2020, pp.
  11\,618--11\,628.

\bibitem{gurobi}
\BIBentryALTinterwordspacing
{Gurobi Optimization, LLC}, ``{Gurobi Optimizer Reference Manual},'' 2023.
  [Online]. Available: \url{https://www.gurobi.com}
\BIBentrySTDinterwordspacing

\bibitem{bernardinEvaluatingMultipleObject2008}
K.~Bernardin and R.~Stiefelhagen, ``Evaluating {{Multiple Object Tracking
  Performance}}: {{The CLEAR MOT Metrics}},'' \emph{EURASIP Journal on Image
  and Video Processing}, vol. 2008, pp. 1--10, 2008.

\bibitem{chiuProbabilistic3DMultiObject2020}
H.-k. Chiu, A.~Prioletti, J.~Li, and J.~Bohg, ``Probabilistic {{3D Multi-Object
  Tracking}} for {{Autonomous Driving}},'' \emph{arXiv:2001.05673 [cs]}, Jan.
  2020.

\end{thebibliography}

\end{document}